\documentclass[letterpaper, 10 pt, conference]{ieeeconf}
\IEEEoverridecommandlockouts                           

\usepackage{graphicx}
\usepackage{soul}
\usepackage{multirow}
\usepackage[english]{babel}
\usepackage[utf8]{inputenc}
\usepackage{xcolor}
\usepackage{longtable}
\usepackage{pdfpages,epstopdf}
\usepackage{cite,esvect}
\usepackage{longtable}
\usepackage{hyperref}
\usepackage{graphicx}
\usepackage[font=footnotesize,labelfont=bf,skip=0pt]{caption}
\usepackage{epstopdf}
\usepackage{amssymb}
\usepackage{amsmath} 
\usepackage{enumerate}
\usepackage{etaremune}
\usepackage{setspace}
\usepackage{comment}
\usepackage{color}
\usepackage{wrapfig}
\usepackage[algo2e]{algorithm2e} 
\usepackage{algorithm} 
\usepackage{algpseudocode} 
\usepackage{cite}
\usepackage{booktabs}
\usepackage{float}
\usepackage{pgfplots} 
\pgfplotsset{compat=newest} 
\pgfplotsset{plot coordinates/math parser=false} 
\newlength\figureheight 
\newlength\figurewidth 
\usetikzlibrary{mindmap}
\usepackage{placeins}
\usetikzlibrary{arrows}
\usepgfplotslibrary{patchplots}

\usepackage{subcaption}

\def\subparagraph{} 
\usepackage{titlesec}
\titlespacing{\subsection}{0pt}{*1}{*1}

\renewcommand{\thesubsubsection}{\arabic{subsubsection}}

\titleformat{\subsubsection}[runin]{\itshape}{\thesubsubsection)}{1em}{}
\titlespacing*{\subsubsection}{\parindent}{0pt}{*1}

\makeatletter
\let\oldlt\longtable
\let\endoldlt\endlongtable
\def\longtable{\@ifnextchar[\longtable@i \longtable@ii}
\def\longtable@i[#1]{\begin{figure}[t]
		\onecolumn
		\begin{minipage}{0.5\textwidth}
			\oldlt[#1]
		}
		\def\longtable@ii{\begin{figure}[t]
				\onecolumn
				\begin{minipage}{0.5\textwidth}
					\oldlt
				}
				\def\endlongtable{\endoldlt
				\end{minipage}
				\twocolumn
		\end{figure}}
		\makeatother

\title{\LARGE \bf
Grasping Force Control and Adaptation for a Cable-Driven\\ Robotic Hand 

\author{Eric Mountain$^{1}$, Ean Weise$^{1}$, Sibo Tian$^2$, Beiwen Li$^3$, Xiao Liang$^{4,*}$, and Minghui Zheng$^{2,*}$
    \thanks{This work was partially supported by the USA National Science Foundation under Grant No. 2026533/2422826 and 2132923/2422640. The authors would also like to recognize Xingsheng Wei for his development of the Integrated-Finger Robotic Hand, Chris Stein for his hardware/software assistance, and Wansong Liu for his intellectual insights.}
    \thanks{$^{1}$ Eric Mountain and Ean Weise are with the Department of Mechanical and Aerospace Engineering, University at Buffalo, Buffalo, NY14260, USA. {\tt\small Emails:\{ericmoun, eanweise\}@buffalo.edu}.}
    \thanks{$^{2}$ Sibo Tian and Minghui Zheng are with the J. Mike Walker '66 Department of Mechanical Engineering, Texas A\&M University, College Station, TX 77840, USA. {\tt\small Emails: \{sibotian, mhzheng\}@tamu.edu}.}
    \thanks{$^{3}$ Beiwen Li is with the Department of Mechanical Engineering, Iowa State University, Ames, IA 50011, USA. {\tt\small Email: beiwen@iastate.edu}.}
    \thanks{$^{4}$ Xiao Liang is with the Department of Civil \& Environmental Engineering, Texas A\&M University, College Station, TX 77840, USA. {\tt\small Email: xliang@tamu.edu}.}
    \thanks{$^*$ Corresponding Authors.}
    }}

\begin{document}
\maketitle
\begin{abstract}
This paper introduces a unique force control and adaptation algorithm for a lightweight and low-complexity five-fingered robotic hand, namely an Integrated-Finger Robotic Hand (IFRH). The force control and adaptation algorithm is intuitive to design, easy to implement, and improves the grasping functionality through feedforward adaptation automatically. Specifically, we have extended Youla-parameterization which is traditionally used in feedback controller design into a feedforward iterative learning control algorithm (ILC). The uniqueness of such an extension is that both the feedback and feedforward controllers are parameterized over one unified design parameter which can be easily customized based on the desired closed-loop performance. While Youla-parameterization and ILC have been explored in the past on various applications, our unique parameterization and computational methods make the design intuitive and easy to implement. This provides both robust and adaptive learning capabilities, and our application rivals the complexity of many robotic hand control systems. Extensive experimental tests have been conducted to validate the effectiveness of our method. 
\end{abstract}

\section{Introduction and Related Work}
\begin{figure*}[t]
	\centering
	\vspace{10pt}
	\includegraphics[scale=0.85]{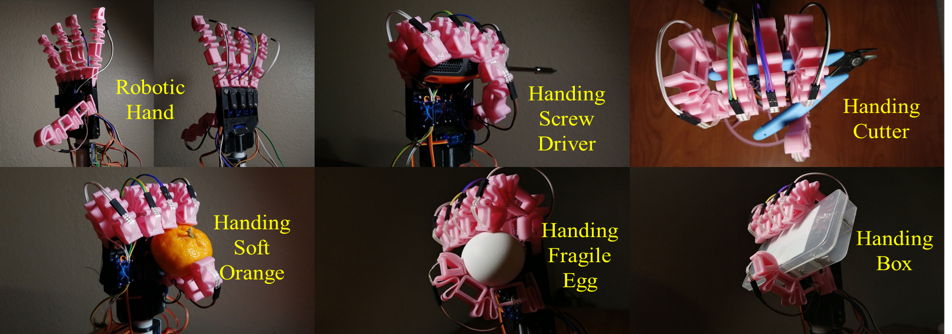}
	\vspace{5pt}
	\caption{Overview of IFRH \cite{wei2022novel}}\vspace{-20pt}
	\label{IFRH}
\end{figure*}

This paper presents the design of an Integrated-Finger Robotic Hand (IFRH, as shown in Figure \ref{IFRH}) and a new grasping force adaptation algorithm to automatically improve its grasping performance via learning over iterations. The 3D-printed structure was designed with cost and weight in mind. The structure and its capability to adjust grasping force make it a viable option for the grasping of a wide variety of objects in many applications such as food processing, agriculture, and disassembly \cite{lee2024review,liu2023task,lee2022task,lee2022robot}.

Grasping force control has been thoroughly studied in recent years within the robotics field, with the first appearances of adaptive controllers debuting commercially in the 1980s \cite{Astrom1989}. One control method researched by Li et al. \cite{Li2014} is grasp adaptation through an object level impedance controller and grasp stability estimator, tested on the Allegro hand. Depending on the tactile feedback data, grasp stiffness and rest length would be adapted within the impedance controllers. The values for grasp stiffness and rest length were chosen based on the training of the grasp stability estimator. While the weight range of graspable objects is increased using this method, a relatively large amount of data is required to train the grasp stability estimator. An additional control method is a normal force-derivative feedback control method applied to the Motion Control Hand, introduced by Engeberg and Meek \cite{Engeberg2008}. This method obtains the normal force derivative by differentiating the normal force measured at the fingertips. This method was compared to proportional force control on the same robotic hand with the same initial conditions and control gains. Experimental results concluded that the normal force-derivative feedback control had increased damping of the closed-loop system as well as reduced overshoot. A downside to this method is that it is more computationally heavy, requiring more code for implementation. A common adaptive control method is model predictive control, which was discussed by Flores et al. \cite{Flores2022} through the application on a piezoelectric actuated robotic hand. The controller uses a nonlinear observer and was modeled using a nonlinear Bouc-Wen model. Experimental results show stabilization of the closed-loop output. A downside to this method is its computational complexity in determining the optimal control action, especially with high-dimension systems.

Our paper leverages the Youla-parameterization and iterative learning control (ILC) to improve the grasping functionality over iterations. When implementing the ILC algorithm, the control signal for each iteration is updated as a function of the previous system error, thereby eliminating the need to design an optimal controller. Studies of iterative learning control in robotics include the discussion of quadratic-criterion-based iterative learning control (Q-ILC) for trajectory tracking of robotic arms by Zhu et al. \cite{Zhu2019}. The Q-ILC method aims to reduce the position error by incorporating the velocity of the robot system. Furthermore, Zhao et al. \cite{Zhao2017} apply iterative learning control to the path-tracking of a mobile robot using feedback-aided p-type ILC. This method incorporates an initial rectifying term to the ILC algorithm to quickly converge the tracking error to zero. Wang et al. \cite{wang2016robust,wang2018robust} utilize robust control theory to design the learning filter by transferring the feedforward control problem into a feedback control problem, which Zheng et al. extended to high-order ILC by transforming the learning filter design problem into an optimal feedback controller design problem \cite{zheng2017design}. Additionally, Xu et al \cite{Xu2020} study an adaptive iterative learning control based on extended state observer in the path tracking of a double-jointed robot. The extended state observer recognizes the internal interference and the external disturbance of the system. Modi et al \cite{modi2024improving} extended iterative learning control to heterogeneous systems. There are also some studies on combining Youla-parameterization and iteration learning control, such as \cite{verwoerd2006admissible,peng2017constrained,afshar2007robust}; however, most of the applications are not in robotics-related areas. More importantly, most of the proposed design methods, either ILC itself or its combination with Youla-parameterization, are not intuitive and require many design experiences. 

This paper presents a systematic feedback and feed-forward controller design procedure for the cable-driven robotic hand. Mathematically, we extend Youla-parameterization that is traditionally used in feedback controller design into a feedforward ILC algorithm. The uniqueness of such extension is that both the feedback and feedforward controllers are parameterized over one unified design parameter which can be easily customized based on the desired closed-loop performance. 

This paper is structured as follows.
Section II will introduce the mechanical design and actuation of the Integrated-Finger Robotic Hand. Section III will discuss the grasping force control and adaptation method, which uniquely parameterizes both the feedback controller and the feedforward iterative learning controller over one parameter. This parameter just needs to be stable and can be designed easily and intuitively based on the desired closed-loop system. Section IV will then present experimental validation of the learning control algorithm and Section V will conclude the paper. 

\section{Brief Introduction of Design and Actuation}

This section will briefly introduce the mechanical design and actuation system of the robotic hand. A more detailed introduction regarding the robotic hand design was presented in the group's prior work \cite{wei2022novel}. The Integrated-Finger Robotic Hand is designed with simplicity and ease of use in mind. The 3D-printed structure is created to mimic a real human hand, with five fingers and 14 knuckles. Each finger on the IFRH has three knuckles, with the exception of the thumb which has two. The 3D-printed material used is Polylactic Acid (PLA) plastic. PLA has many benefits, including its low cost, low warping tendencies, print detail, and versitility. The low weight of the IFRH design does not prevent the hand from being able to grasp daily objects, as demonstrated in Figure \ref{IFRH}, reflecting the wide range of shapes, fragility, and weight of objects that the IFRH can clasp. An added benefit to this design is that if one of the 3D-printed components were to break, the user can quickly and cheaply replace the part. This is in contrast to many commercial robotic hands which are expensive or tedious to fix. 

\begin{figure}[!htbp]
	\centering
	\includegraphics[scale=0.58]{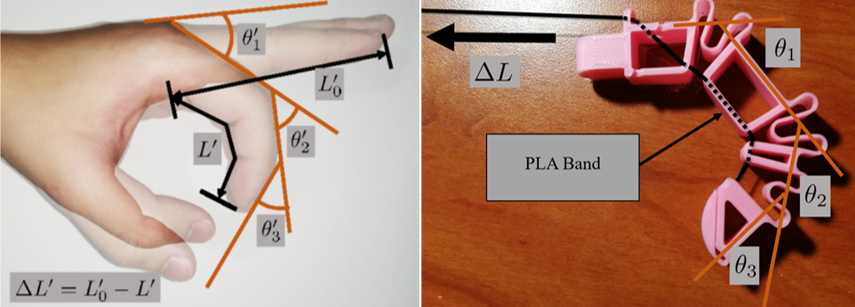}
	\vspace{5pt}
	\caption{Transmission of IFRH compared with a human hand \cite{wei2022novel} ($\theta_{1,2,3}$: angles of EKC; $\Delta L$: pulled distance of PLA Band; $\theta'_{1,2,3}$: angles of human finger joint;$\Delta L'$: length changes of human finger during bending.)}
	\label{Robotic Finger Bend}
\end{figure}

In Figure \ref{IFRH}, one can see the imitated knuckles on the fingers, known as the Elastic Knuckle Connections (EKC). These are used to store elastic energy during the finger's bending motion. Tiankoncng servo motors with a weight of 9g are attached to the base of all fingers, except the thumb which uses a 55g Tiankoncng servo motor. The larger Tiankoncng motor gives the thumb more torque when grasping objects, which is needed to counteract the opposing forces of the other four fingers as well as the force of gravity. A thin band of PLA runs through each of the fingers, connecting to the corresponding servo motor on one end, and to the fingertip on the other. When the servo motor turns, the thin band is contracted, causing the finger to simulate a closing motion. This process is visualized in Figure \ref{Robotic Finger Bend}.

\begin{figure*}
	\centering
	\includegraphics[scale=0.6]{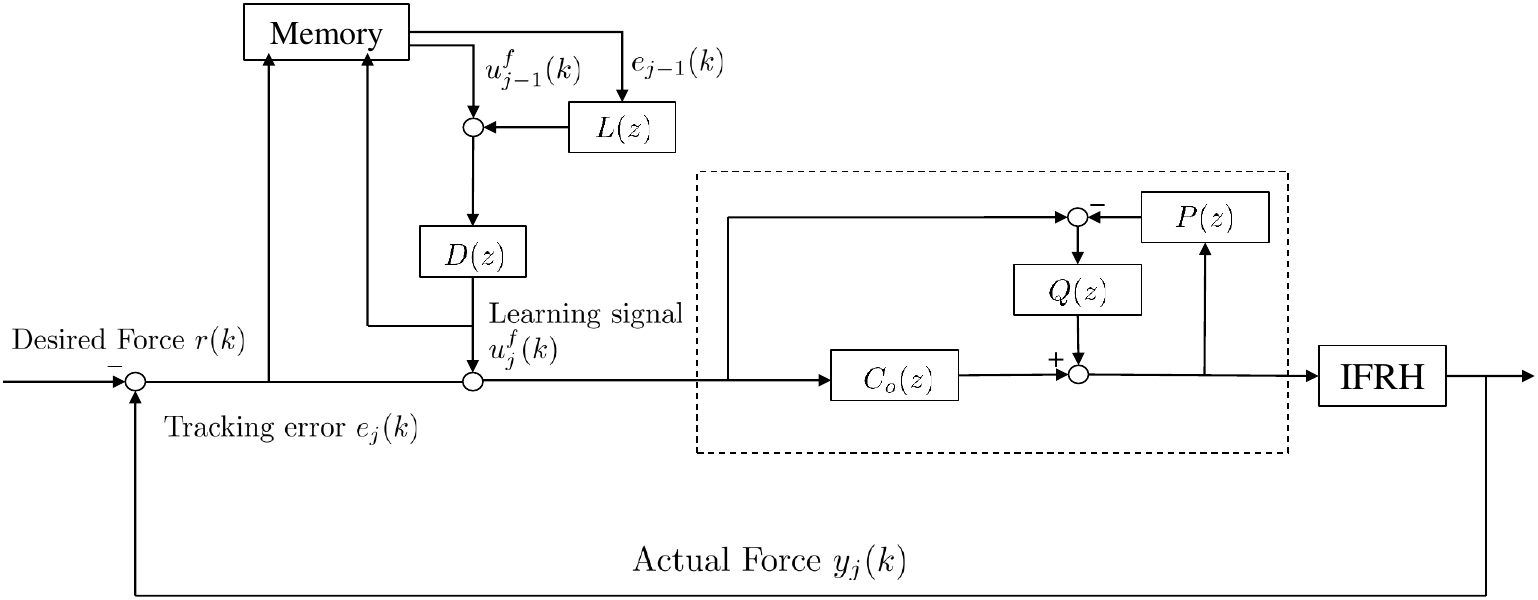}
	\vspace{5pt}
	\caption{Grasping Adaptation Framework using ILC}
	\label{OverallControl}
\end{figure*}

Along with the finger's actuation, we can also see the defining angles and changes in finger length that contribute to the tracking of the IFRH finger's position. We can compare the actuation of the IFRH finger with the bending motion of a human finger using these variables. The bending angles and positions for a human hand are defined in Figure \ref{Robotic Finger Bend}. The apostrophes in Figure \ref{Robotic Finger Bend} are present to denote the measurements of the human finger, whereas the measurements of the IFRH finger do not have apostrophes. Located within each fingertip is a Force Sensitive Resistor (FSR). The reading sent from the FSR is a resistance, which can then be converted to a force in Newtons. The measurement range of a single Force Sensitive Resistor is between 0 and 19.62 Newtons, with a sensing area of $40 \, \mathrm{mm}^2$. When the sensor comes into contact with an object, we can measure the output force on each finger. This allows us to create a force-dependent control system. The benefit of this over a time-dependent control system is that the reference force can dictate the grasping control of the hand, rather than a preset time which varies often with differently sized objects.  

In summary, the IFRH's low complexity, low weight, low cost, and high versatility enable its viability within a wide range of applications and simplify the design of controller and force adaptation.

\section{Grasping Force Control and Adaptation}

This section will introduce the overall control and adaptation algorithm to automatically adjust the grasping force. It consists of a Youla-parameterization based feedback controller and a feedforward iterative adaptation controller. Youla-parameterization is beneficial due to its flexibility in controller design, simplified control structure, and tunable parameters, making it a useful tool when the order of the controller is free. 
Due to modeling uncertainties and the limitation of the feedback control, we design an adaptation algorithm for the grasping force based on iterative learning control. This further improves the grasping performance over iterations. In brief, we leverage both Youla-parameterization and ILC in our overall control framework, and synthesize the feedback control and feedforward control design using the Youla parameter $Q(z)$ which can be any stable proper system and can be designed easily based on a desired close-loop performance.

The overview of the control structure is provided in Figure \ref{OverallControl}, where we define $r(k)$ as the reference input, $e(k)$ as the system error, $C_O(z)$ as the baseline feedback controller, $P(z)$ as the system plant, $u(k)$ as the control signal, $y(k)$ as the output, $u^f(k)$ as the learning control signal, and the "Memory" as the stored error per iteration. There are three design parameters, $D(z)$ as a low-pass filter, $L(z)$ as a learning filter, and $Q(z)$ as a Youla parameter. In this section, we will introduce details on how to design the three parameters.

The stabilizing controller $C(z)$ can be parameterized in the way shown in the dotted box of Figure \ref{OverallControl}, from which the following relationship can be obtained
\begin{equation}
	C(z) = \frac{C_o(z)+Q(z)}{1 - P(z)Q(z)}
	\label{Cz}
\end{equation}
where $Q(z)$ can be a stable proper system that is to be designed. In addition to the tracking error signal $e(k)$ sent to the controller $C(z)$, we add a learning signal $u^f(k)$ which is generated from the last iteration information, including the learning signal and tracking errors from the last iteration. Here we denote the current iteration as $j$ and the previous one as $j-1$. This learning signal aims to adjust the grasping force performance over iterations in a repetitive mode. The system's performance is measured by tracking the difference between the system's output and the reference input, where the reference input remains unchanged through each iteration. The reference input for our system is the desired target force of the thumb contacting the object, and the system output is the force sensor reading from the thumb Force Sensitive Resistor. The error is measured as an accumulation of discrete values within a stored array. This array and the reference input are used to update the control signal for the next iteration. In the following description, we omit the z-variable to make the equations neat.

In this paper, we design the grasping force adaptation law as follows
\begin{equation}
u_{j}^f = D(u_{j-1}^f+Le_{j-1})
\label{uj1}
\end{equation}
where $D$ is a low-pass filter and $L$ is a learning filter. In the following paragraphs, we will derive the relationship between the tracking error of the current iteration and that of the previous iteration, i.e., the relationship between $e_{j}$ and $e_{j-1}$. Considering the desired force is a low-frequency signal and the modeling uncertainties happen at relatively high frequency, $D$ is designed as a low-pass filter here to secure some robustness to system uncertainties without significantly sacrificing the tracking performance. To simplify the derivation, we can approximately treat $D$ as 1. Therefore, at the $j^\text{th}$ iteration, the error for the current iteration $e_{j}$ can be obtained by passing the previous iteration's error $e_{j-1}$ through the closed-loop transfer function $T_u$ as well as the learning filter $L$ and then adding the result to the current iteration's error, i.e.,
\begin{equation}
	e_j = T_uu_j^f + T_rr
	\label{ej}
\end{equation}
where $T_u$ represents the closed-loop transfer function from the learning signal $u^f$ to the tracking error $e$ and $T_r$ represents the closed-loop transfer function from the reference $r$ to the tracking error $e$. These closed-loop transfer functions can be found using the equations below, where $C$ is the controller design within the dotted lines, and P is the IFRH system plant.
\begin{equation}
	T_u = -(1+PC)^{-1}PC
	\label{Tu}
\end{equation}
\begin{equation}
	T_r = (1+PC)^{-1}
	\label{Tr}
\end{equation}
Plugging into the adaptation law in Equation (\ref{uj1}) by assuming $D(z)$ as 1, we have
\begin{equation}
	e_j = T_u(u_{j-1}^f+Le_{j-1}) + T_rr
\end{equation}
Considering the $j-1$ iteration, where
\begin{equation}
	e_{j-1} = T_uu_{j-1}^f + T_rr
\end{equation}
we can have
\begin{equation}
	e_j = e_{j-1}-T_rr+T_uLe_{j-1} + T_rr
\end{equation}
which further results in
\begin{equation}
e_{j+1} = (1+T_uL)e_j
\label{ej1}
\end{equation}

Now we represent $T_u$ in terms of $Q(z)$. From Equations (\ref{Cz}) and (\ref{Tu}), we have
\begin{equation}
	T_u = -\frac{P(C_o+Q)}{1+PC_o}
\end{equation}
To minimize $e_j$, based on the Equation (\ref{ej1}), we can select $L(z)$ as the negative inverse of the closed-loop transfer function from the learning signal to the tracking error, i.e., $L(z)$ can be designed as follows:
\begin{equation}
L = \frac{1+PC_o}{P(C_o+Q)}
\label{L}
\end{equation}
Until now, we parameterize both the learning filter and the controller by the design parameter $Q(z)$, as summarized in Table \ref{DLcl}.

\begin{table}[ht]
	\centering
	\begin{tabular}{lll}
		\toprule[1.5pt]		Items  & Design \\\toprule[1.5pt]
		$D(z)$ & Low-pass filter\\\vspace{0.2cm}
		$Q(z)$ & Any stable proper system\\\vspace{0.2cm}
		$L(z)$ & 
			$L(z) = \frac{1+PC_o}{P(C_o+Q)}$
\\\vspace{0.2cm}
		$C(z)$ &  
			$C(z) = \frac{C_o+Q}{1 - PQ}$
	\\\toprule[1.5pt]
	\end{tabular}
	\vspace{5pt}
	\caption{Design Parameterization}
	\label{DLcl}
\end{table} Ideally, if $L(z)$ is designed by Equation (\ref{L}), this minimizes the norm of $1+T_uL$ and thus optimizes the learning performance. The only requirement is that $L(z)$ is stable for any stable $Q(z)$. The only missing information now is how to choose $Q(z)$ to guarantee a closed-loop performance. 
The closed-loop transfer function $G_c(z)$ from the reference $r(k)$ to the actual force $y(k)$ can be obtained as follows
\begin{equation}
	G_c=\frac{PC}{1+PC}=\cfrac{P\cfrac{C_o+Q}{1 - PQ}}{1+P\cfrac{C_o+Q}{1 - PQ}}=
	\cfrac{P(C_o+Q)}{1+PC_o}\label{Gc}
\end{equation}
which is an explicit relationship between the closed-loop transfer function $G_c(z)$ and $Q(z)$. This relation provides us an easy way to design $Q(z)$ based on $G_c(z)$, and from which the controller $C(z)$ and the learning filter $L(z)$ can be obtained immediately according to Table \ref{DLcl}.

\section{Experimental Validation}

The experimental studies consist of three subsections: system identification, Youla-parameterized controller design, and grasping force adaptation. 
\begin{figure}[!htbp]
	\centering
	\includegraphics[scale=0.35]{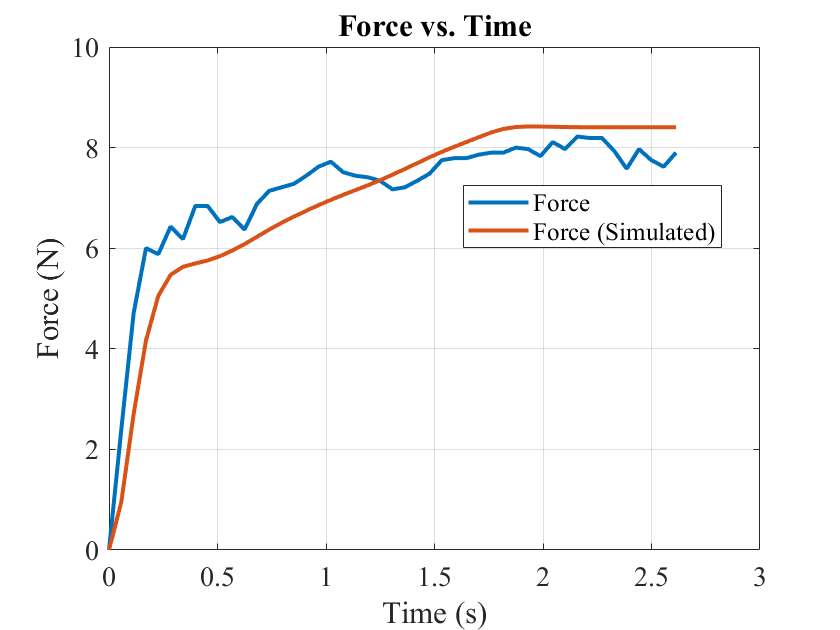}\vspace{5pt}
	\caption{System Identification: Simulated Output Using Identified Model vs. Actual Experimental Output Data}
	\label{Sim_Actual}
\end{figure}

\begin{figure}[!htbp]
	\centering
	\includegraphics[scale=0.45]{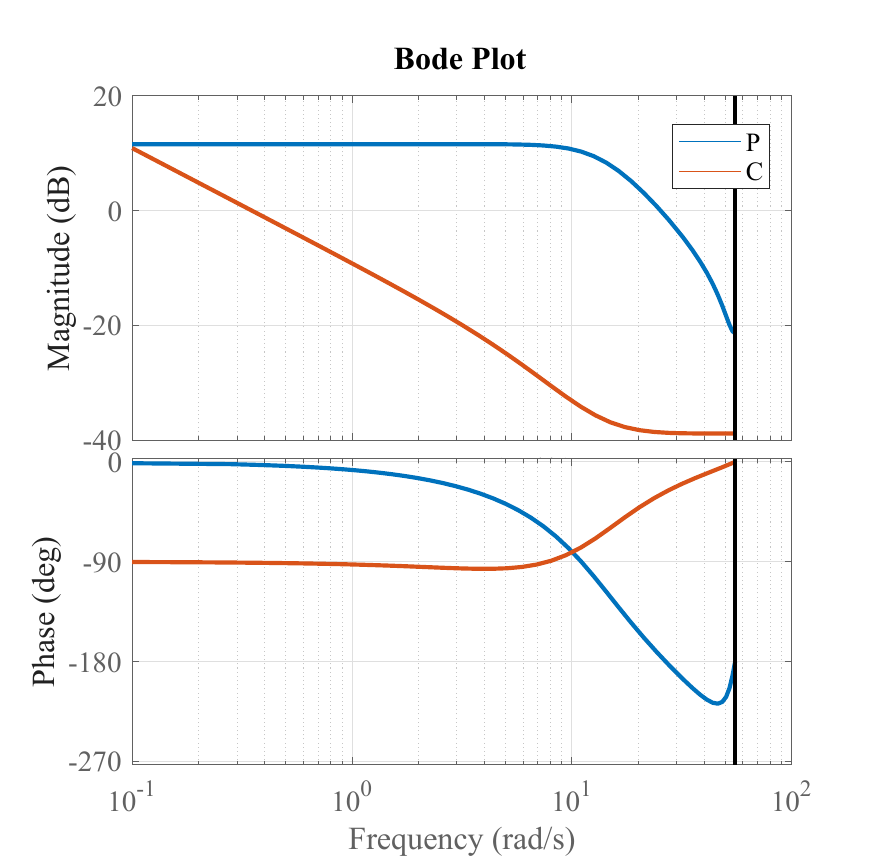}
	\caption{Bode Plots for the Controller and Plant}
	\label{Bode-1}
\end{figure}

\subsection{System Identification}
We first introduce the method of system identification for the Integrated-Finger Robotic Hand using the input and output data of the robotic hand. The input to the system is the position of the finger's servo motor during the grasping motion. The output of the system is the force measured by the finger's force sensitive resistor. The system was identified using experimental data of the thumb grasping a spool of rigid thread. This spool was chosen based on its 2.25 inch diameter, allowing for comfortable grasping by the thumb. An experiment was conducted of the grasping of this spool until steady-state conditions were met. Data of the servo motor position and force readings over time were collected.

The data was collected via Arduino, with a sampling rate of 0.0568 seconds. The data is modeled from the start of the grasping process, when the finger comes into contact with the object. Furthermore, in the experiment described above, a target force of 10 Newtons was set to grasp the spool of thread. This value was determined experimentally. Based on the force data in Figure \ref{Sim_Actual}, we modeled the open-loop plant as a second order system. 

\begin{equation}
P(z) = \frac{b_1z + b_2}{z^2 + a_1z + a_2}
\label{Pz1}
\end{equation}

The coefficients are found using the experimental data collected and the recursive least square estimation method.  Our system transfer function is identified as follows:

\begin{equation}
P(z) = \frac{0.7902z + 0.6208}{z^2 - 0.9748z + 0.3442}
\label{Pz2}
\end{equation}

The simulation of the derived system output compared to real-world force sensor data can be seen in Figure \ref{Sim_Actual}. {The same system identification process can be applied to the other fingers to model their dynamics.}

\subsection{Integrated Design of Controller and Learning Filter}

Since our plant is stable from this method of system identification, $C_o(z)$ in Figure \ref{OverallControl} can be set equal to zero. Also, we chose the desired closed-loop transfer function in continuous-time domain as follows
\begin{equation}
G_{c,\text{continuous}}(s) = \frac{9}{s^2 +3s + 9}
\end{equation}
where the poles are located at -3, -3. This was chosen so the system has a practical settling time (2.67 seconds). The corresponding discrete-time transfer function can be obtained as follows 
\begin{equation}
G_c(z)= \frac{0.013z + 0.0116}{z^3 - 1.687z^2 + 0.711z}
\end{equation}
Plugging it into Equation (\ref{Gc}), we obtain $Q(z)$ as follows:
\begin{equation}
Q(z) = \frac{0.013z^4 - 0.0012z^3 - 0.0069z^2 + 0.004z}{0.784z^4 - 0.737z^3 + 0.430z^2 +0.416z}
\end{equation}
which will be used to obtain the controller and the learning filter based on Equations \ref{Cz} and \ref{L}:
\begin{equation}
C(z) = \frac{0.017z^2 - 0.0159z + 0.0053}{z^2 - 1.676z + 0.676}
\end{equation}
and \begin{equation}
	L = \frac{14.98z^5 {-} 39.84z^4 {+} 40.18z^3 {-} 18.65z^2 {+} 3.413z {+} 0.0467}{z^4 {-} 0.1865z^3 {-} 0.3842z^2 {+} 0.2336z}
\end{equation}
The bode plots for the plant $P(z)$ and Controller $C(z)$ are provided in Figure \ref{Bode-1}.

\subsection{Grasping Adaptation Algorithm Design and Verification}

\begin{figure}[!htbp]
	\centering
	\includegraphics[scale=0.25]{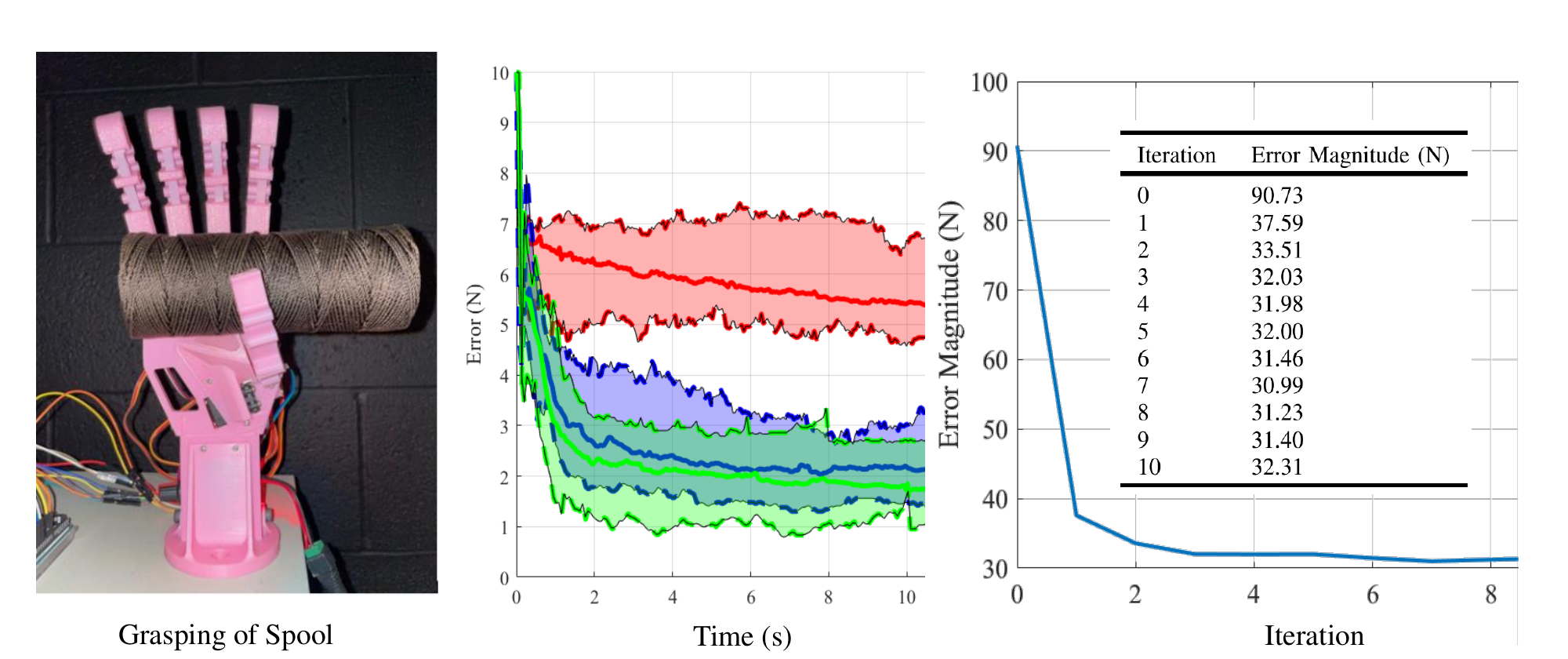} \vspace{5pt}
	\caption{Grasping of Spool of rigid thread. Left: Experimental Scenario. Middle: Average errors over 10 experiments in time domain for the first two iterations (purple and green respectively) and the one without iteration (red). Right: Average error norm over iterations (including iteration 0). 
	}
	\label{fig:Spool}
\end{figure}
\begin{figure}[!htbp]
	\centering
	\includegraphics[scale=0.25]{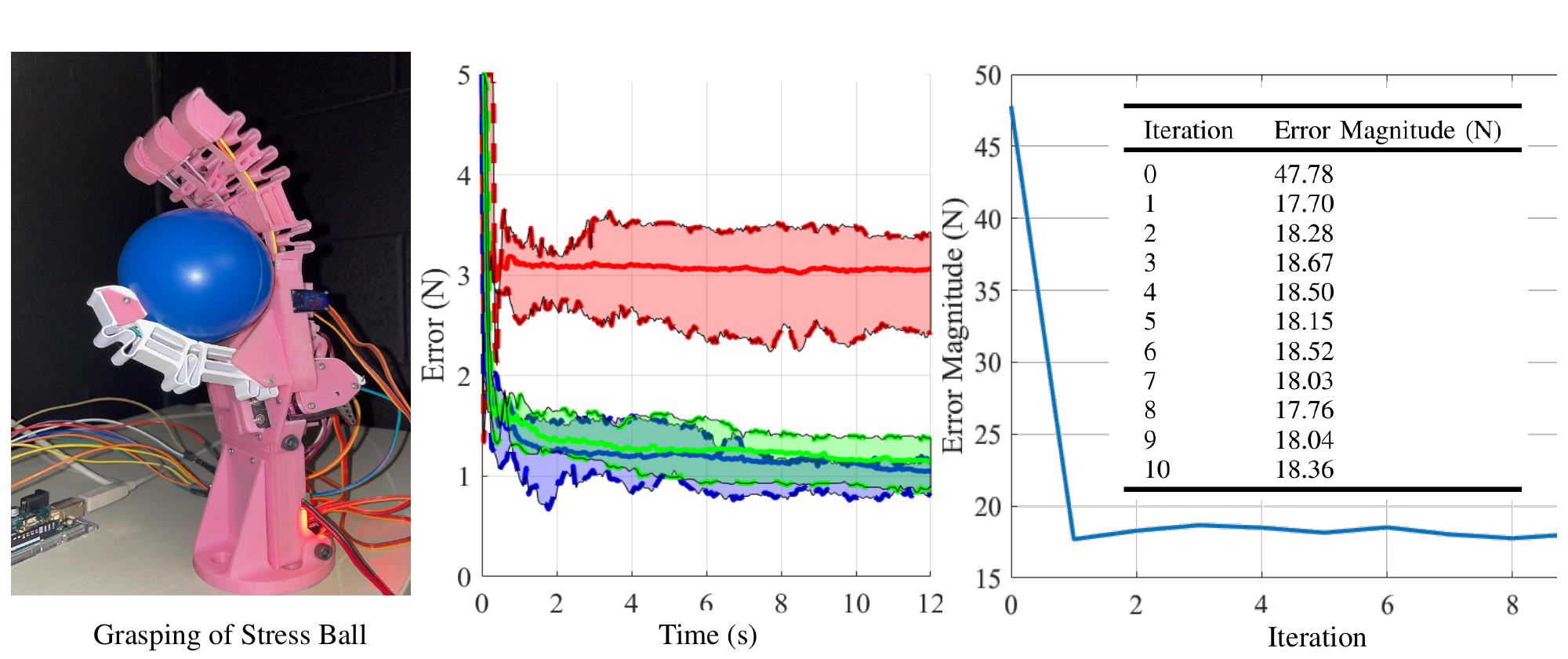}\vspace{5pt}
	\caption{Grasping of Stress Ball. Left: Experimental Scenario. Middle: Average errors over 10 experiments in time domain for the first two iterations (purple and green respectively) and the one without iteration (red). Right: Average error norm over iterations (including iteration 0).}
	\label{fig:Ball}
\end{figure}
\begin{figure}[!htbp]
	\centering
	\includegraphics[scale=0.25]{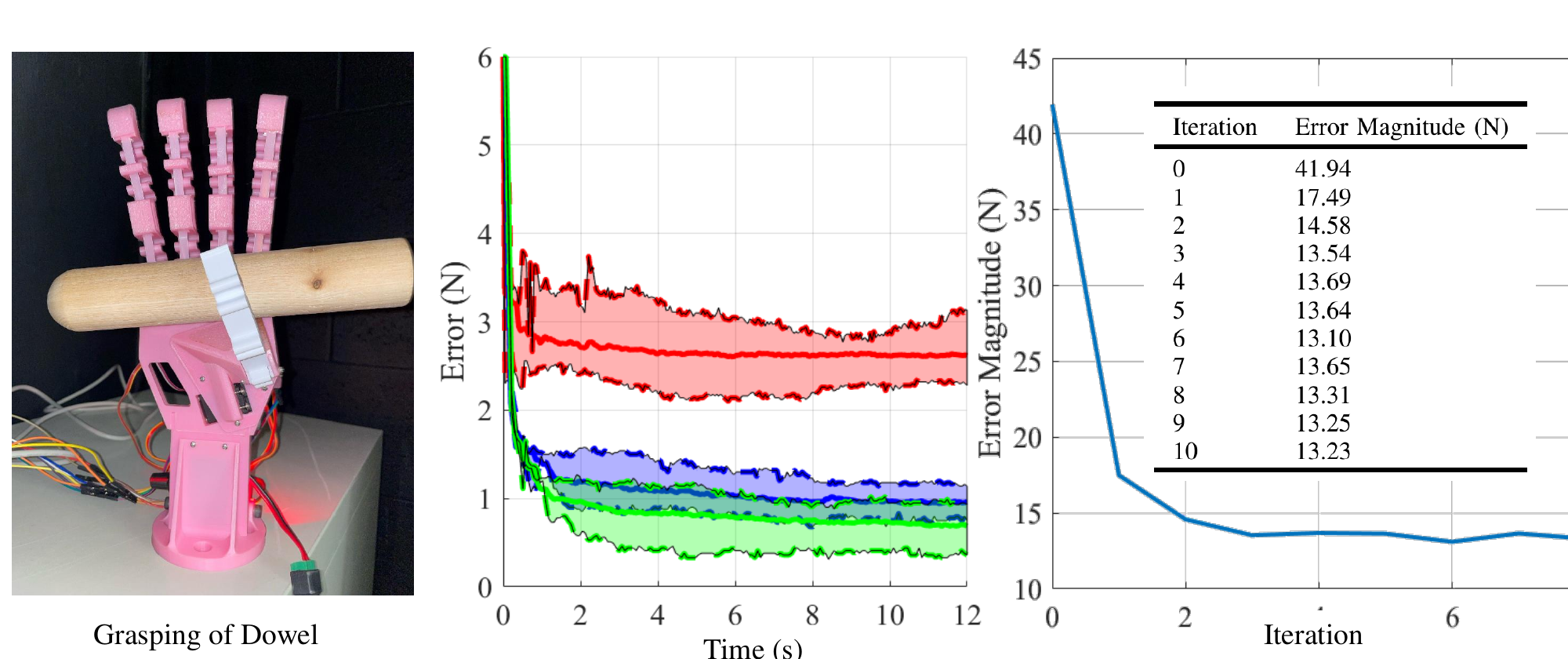}\vspace{5pt}
	\caption{Grasping of Dowel. Left: Experimental Scenario. Middle: Average errors over 10 experiments in time domain for the first two iterations (purple and green respectively) and the one without iteration (red). Right: Average error norm over iterations (including iteration 0).}
	\label{fig:Dowel}
\end{figure}

We have tested our proposed algorithm on three different objects: a spool of rigid thread, a stress ball, and a wooden dowel, as shown in Figure \ref{fig:Spool}, Figure \ref{fig:Ball} and Figure \ref{fig:Dowel}. The stress ball is chosen for its contrast to the spool of thread, given it has a more circular shape and requires a lower grasping force due to its lower rigidity, and the wooden dowel is chosen for its rigidness. The experiment is conducted 10 times for each object, and the averages are analyzed for consistency in the results. When the iteration 0 code is uploaded to the robotic hand, the error data $e_0$ is stored such that it can be used in the following iteration. Utilizing Equation (\ref{uj1}) and our values for $D(z)$, $L(z)$, and $e_0$ we can solve for our first updated control signal, $u_1^f$. This new control signal will be used in the next iteration, exemplifying the system feedback loop in Figure \ref{OverallControl}. Through the next iteration, another error vector will be collected, and the process repeats until a convergence criteria is reached. 
 The experimental results for the spool of rigid thread are provided in Figure \ref{fig:Spool}. The left image shows the experimental scenario. The average error norm of each iteration can be seen in the right figure, which shows that iteration 0 has the largest magnitude of error, then a large drop to iteration 1, and another two small improvements for iterations 2 and 3. Beyond iteration 3, there is no discernible trend in the system error. For a better visualization of the improvements made within the first three iterations, iterations 0, 1, and 2 are plotted in the middle figure. Both the minimum and maximum errors over time for each of these iterations is displayed, taken from the 10 experimental trials. Additionally, the average error norm for each iteration is presented in the rightmost plot in Figure \ref{fig:Spool}. This displays a clear improvement in accuracy once the learning control is applied. It is noted that the error over time does not reach zero, which can likely be attributed to the sensor accuracy and servo motor limitations. 
 We also tested the robotic hand with the proposed grasping adaptation algorithm on a stress ball and a wooden dowel. These experiments show similar results, i.e., the grasping force can be automatically adjust via learning from previous iterations and the performance has been improved effectively, as shown in Figures \ref{fig:Ball} and \ref{fig:Dowel}.

\section{CONCLUSIONS}

This paper introduced a unique force control and adaptation algorithm based on Youla-parameterization and iterative learning control for a lightweight and low-complexity five-fingered robotic hand, namely an Integrated-Finger Robotic Hand (IFRH). The force control and adaptation algorithm is intuitive to design, easy to implement, and improves the grasping functionality through feedforward adaptation automatically. 
Through the implementation of this control strategy, we concluded that updating our control signal with each iteration reduced the overall system error. We further found that the error converged after two-three iterations for each experiment, demonstrating the effectiveness of the control and adaptation algorithm. The successful experimental results conclude that our proposed control strategy works well for the Integrated-Finger Robotic Hand.

\bibliographystyle{unsrt}
\bibliography{references}

\end{document}